\documentclass[10pt,twocolumn,letterpaper]{article}

\usepackage{cvpr}              

\usepackage{graphicx}
\usepackage{amsmath}
\usepackage{amssymb}
\usepackage{booktabs}
\usepackage{enumitem}
\usepackage{caption}
\usepackage{color, colortbl}
\usepackage{float}
\usepackage[caption=false]{subfig}

\usepackage{url}
\newcommand{\tablestyle}[2]{\setlength{\tabcolsep}{#1}\renewcommand{\arraystretch}{#2}\centering\footnotesize}

\definecolor{Gray}{rgb}{0.9,0.9,0.9}
\definecolor{Gray2}{rgb}{0.4,0.4,0.4}

\usepackage[pagebackref,breaklinks,colorlinks]{hyperref}

\usepackage[capitalize]{cleveref}
\crefname{section}{Sec.}{Secs.}
\Crefname{section}{Section}{Sections}
\Crefname{table}{Table}{Tables}
\crefname{table}{Tab.}{Tabs.}

\def\cvprPaperID{6699} 
\def\confName{CVPR}
\def\confYear{2023}

\begin{document}

\title{Rethinking Video ViTs: Sparse Video Tubes for Joint Image and Video Learning}

\author{AJ Piergiovanni \\
\and
Weicheng Kuo \\
\and
Anelia Angelova\\
 }

\maketitle

\begin{abstract}
We present a simple approach which can turn a ViT encoder into an efficient video model, which can seamlessly work with both image and video inputs. By sparsely sampling the inputs, the model is able to do training and inference from both inputs. The model is easily scalable and can be adapted to large-scale pre-trained ViTs without requiring full finetuning. The model achieves SOTA results and the code will be open-sourced.

\end{abstract}

\section{Introduction}
\label{sec:intro}

Visual Transformers (ViT)~\cite{dosovitskiy2020image} have been an ubiquitous backbone for visual representation learning, leading to many advances in image understanding~\cite{scalingVIT,pyramidVIT,segmenter}, multimodal tasks \cite{filip,akbari2021vatt,merlotreserve,flamingo} and self-supervised learning \cite{beit,tong2022videomae,feichtenhofer2022masked}, etc. However, adaptations to video are both challenging and computationally intensive, so video versions have been been specially designed to handle the larger number of frames, for example, ViViT~\cite{arnab2021vivit}, MultiView~\cite{yan2022multiview}, TimeSFormer \cite{bertasius_arxiv_2021} and others~\cite{fan2021multiscale}. 


Video understanding is an essential computer vision task,
and a large number of successful video architectures have been developed~\cite{Carreira_2017_CVPR,tran_iccv_2015,xie_s3d_eccv_2018,wang2018nonlocal,ryoo2020assemblenet++,feichtenhofer_cvpr_2020,feichtenhofer_iccv_2019,liu2021video}.
Previous video 3D CNNs \cite{Carreira_2017_CVPR,tran_iccv_2015} were designed to handle videos by learning spatio-temporal information; they often borrow from mechanisms for learning on images, for example \cite{Carreira_2017_CVPR} use pre-trained image CNN weights by inflating the kernels to 3D. 
However, once adapted to videos, these kernels are no longer applicable to images. 

Furthermore, most previous works treat image and video as entirely different inputs, providing independent methods for either videos or images, since designing a model capable of handling both is challenging. At the same time, image and video inputs are inherently related and a single visual backbone should be able to handle either or both inputs. 
Previous methods for co-training image and video \cite{zhang2021co,likhosherstov2021polyvit,bain2021frozen,omnivl} adapt the architectures to do so with significant portions of the network designed for each input. Works such as Perceiver \cite{jaegle2021perceiver} and Flamingo \cite{flamingo} address this by resampling the input and compressing it into a fixed number of features. However, this resampling can still be expensive for long videos, and, in the case of Flamingo, it treats videos as individual frames sampled at 1 FPS, which limits the temporal information. 
Such low FPS sampling and per-frame modeling would often be insufficient for datasets which rely on motion and temporal understanding, e.g., SomethingSomething \cite{somethingsomething}, 
or for recognizing quick and short actions. On the other hand, using one of the above-mentioned approaches with dense frames is computationally infeasible.

\begin{figure}
    \centering
    \includegraphics[width=1.0\linewidth]{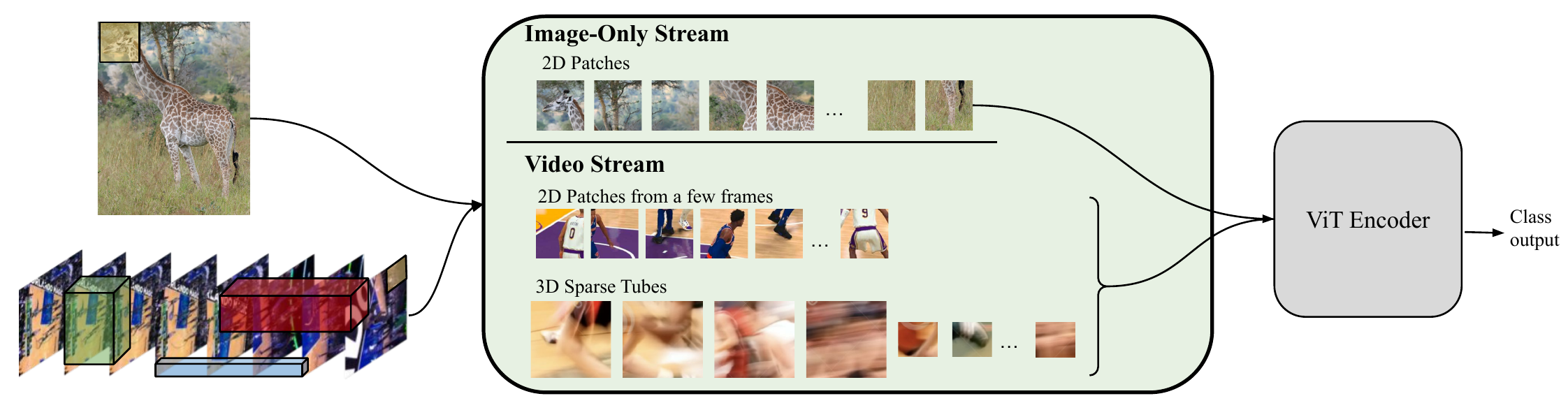}
    \caption{TubeViT: With Sparse Video Tubes, Vision Transformers (ViTs) use both image and video inputs, providing an efficient video backbone and more accurate performance.}
    \label{fig:teaser}
\end{figure}

To address these limitations, we propose a simple but effective 
model, named TubeViT, to utilize a standard ViT model seamlessly for both image and videos. 
We introduce Sparse Video Tubes, a lightweight approach for joint image and video learning.
Our method works by sparsely sampling various sized 3D space-time tubes from the video to generate learnable tokens, which are used by the vision transformer (Figure~\ref{fig:teaser}). With sparse video tubes, the model is easily applicable to either input, and can better leverage either or both sources of data for training and fine-tuning.  
The sparse video tubes naturally handle raw video signals and image signals which is crucial to understanding actions and other spatio-temporal information in videos.

Video models are also expensive to train, and previous works have studied ways to leverage already trained models, such as using frozen ones \cite{lin2022frozen} or adapting them to videos \cite{ni2022expanding}. We expand on these ideas, and use the Sparse Video Tubes to adapt much larger ViT models to videos with lightweight training (Sec. \ref{sec:scaling}). Thus we create powerful large video models with less resources.


We evaluate the approach across many standard video datasets: Kinetics-400, Kinetics-600, Kinetics-700, and SomethingSomething V2, outperforming the state-of-the-art (SOTA). Our methods are trained from scratch or on ImageNet-1k and Kinetics datasets and outperform even methods additionally pre-trained from very large datasets (e.g., JFT~\cite{jft}). Our work also outperforms models targeting video pretraining, 
such as recent video Masked Auto-Encoder (MAE) works~\cite{tong2022videomae,feichtenhofer2022masked}.

Our key findings are that by using the sparse video tubes, we are able to better share the weights learned for both images and videos. This is in contrast to prior works that either inflate kernels or add new temporal-specific layers. Further, due to the sparse sampling, the number of tokens remains low, which we also find is important, both for reducing FLOPs and improving performance.

\textbf{Our contribution} is construction of sparse video tubes, obtained by sparsely sampling videos with various sized 3D space-time tubes. With that we accomplish the following: (1) a universal visual backbone which easily adapts a ViT architecture to videos; (2) joint image and video understanding which seamlessly uses either input; (3) an easy-to-scale approach for video understanding, which can also leverage already trained (large) ViT models.

\section{Related work}
Video understanding is an important topic in computer vision. Early works hand-designed trajectory features to understand motion and time \cite{wang2013action}. With the success of neural networks, many different approaches have been developed, such as two-stream CNNs taking image frames plus optical flow for motion information as input \cite{simonyan_neurips_2014}, finding a clear benefit from adding the flow information. Works studying 3D CNNs found the learning of temporal kernels to be important \cite{tran_iccv_2015,Carreira_2017_CVPR,tgm,tran_cvpr_2018}, but also required much more data in order to be effective \cite{Carreira_2017_CVPR}. Many of the existing video CNN approaches, have been specialized to handle videos, either with flow streams or 3D kernels and thus have not been applicable to images.

With the introduction of transformer models and self-attention \cite{vaswani2017attention}, vision transformers have been very effective for image-based tasks. However, due to the quadratic cost of self-attention and the dense sampling, their use for videos has required different elements, such as space-time factorized attention \cite{bertasius_arxiv_2021, arnab2021vivit,yan2022multiview}. However, these video transformers have not really been tested on longer videos and are mostly evaluated on short 
clips. The ability to handle larger number of input frames and understand long-term actions and their relationships is of key importance, but becomes computationally prohibitive with current models.

\begin{figure}
    \centering
    \includegraphics{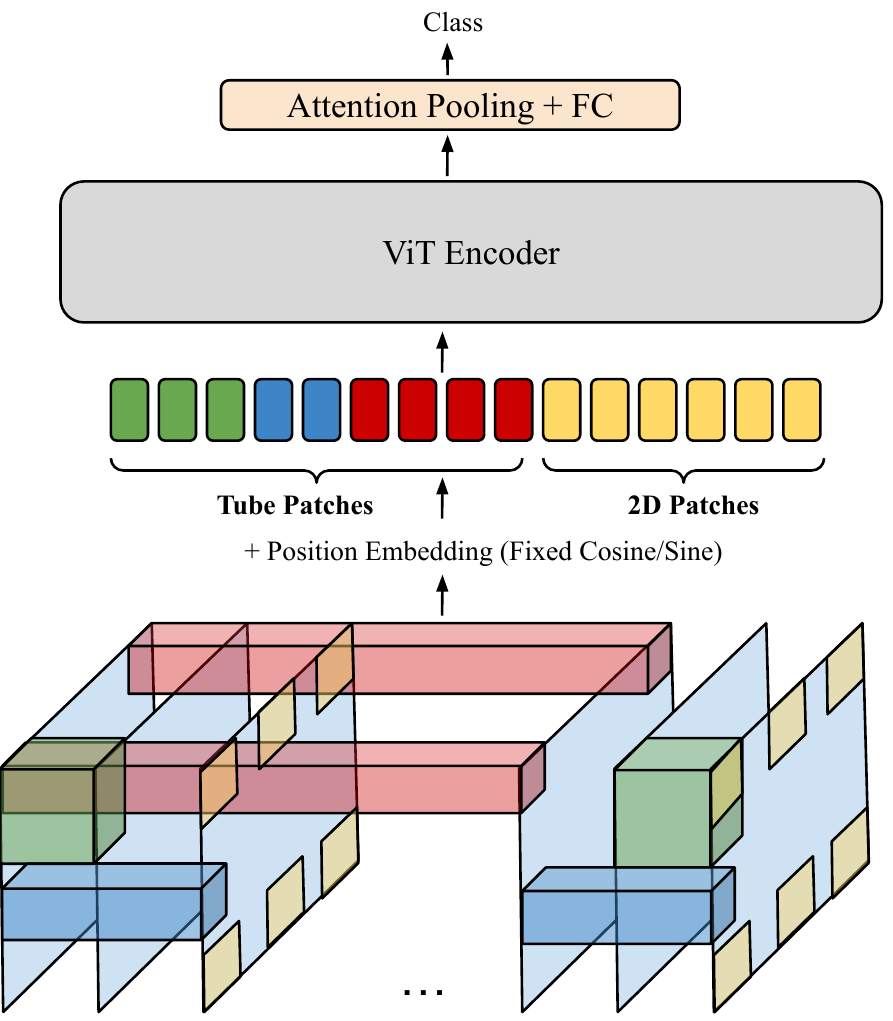}
    \caption{Illustration of the approach. We use tubes of different shapes to sparsely sample the video. These are concatenated together and used as input to a transformer model.}
    \label{fig:sparse_tubes}
\end{figure}

Previous works have found that transformers focus on only a few tokens 
\cite{naseer2021intriguing,rao2021dynamicvit} and works have been designed to pool or reorganized tokens effectively \cite{liang2022notpatches,ryoo2021tokenlearner_neurips,marin2021token}. Many video works have found that frames contain redundant information, and thus propose strategies to sample frames \cite{gowda2021smart,wu2019adaframe}. Other works have studied ways to reduce the number of tokens in video transformer models \cite{wang2022efficient,sparsesampling,ryoo2021tokenlearner_neurips}. However, all these works still use an initial dense sampling of the video, then some heuristics to reduce the number of inputs. In this work, we more sparsely sample the input initially, increasing efficiency. 

Other recent works have studied video MAE tasks as pretraining \cite{feichtenhofer2022masked,tong2022videomae}, they similarly treat videos as tubes, and study the sparseness in terms of the masking, having similar findings that sparseness is beneficial. However, they use a single tube shape and create non-overlapping patches and have not been studied when joint training with images.

This work is also related to approaches which use multiple views or streams from the input data, e.g., Multi-View Transformers \cite{yan2022multiview}, SlowFast Networks \cite{feichtenhofer_iccv_2019} and others \cite{piergiovanni2022cotok,simonyan_neurips_2014}, all have found benefits from multiple input views or streams.
MultiView Transformers \cite{yan2022multiview}, similarly to us, is using tubes of varying shapes. The key difference is the sparse sampling we use enables the use of a single ViT encoder model, rather than multiple smaller, per-view encoders. This further unifies the approach with images.

Another line of work in video understanding is leveraging image datasets during pre-training~\cite{duan2020omni,UniDual}. This is valuable as image-only datasets are better annotated and provide richer semantic information. One approach is to bootstrap the video models from image-pretrained models, often by inflating kernels. The model is first pre-trained on image data, and then only trained on video. Other works proposed to co-train image and video jointly \cite{UniDual,zhang2021co,likhosherstov2021polyvit,bain2021frozen,omnivl,jaegle2021perceiver}. These approaches adapt the architectures to handle both inputs which might be inefficient, e.g., treating an image input as a video of 1 frames~\cite{zhang2021co} or using separate networks to first encode the inputs\cite{jaegle2021perceiver,flamingo}. 


In contrast to all the previous works, our method is simple and straightforward. One crucial set of differences is that the tubes are sparsely applied to the raw input, consists of different shaped, possibly overlapping tubes, and uses a single, shared backbone network, different from all previous approaches (\cite{feichtenhofer_iccv_2019,yan2022multiview,sparsesampling,ryoo2021tokenlearner_neurips,arnab2021vivit,feichtenhofer2022masked,tong2022videomae}). This leads to both more efficient and accurate models. Secondly, and more importantly, the model is entirely shared between the image and video modalities. This is an important distinction as it not only improves performance for both tasks, but is also more generally applicable to vision tasks.

\section{Method}

\subsection{Preliminaries}
The standard ViT architecture \cite{dosovitskiy2020image} takes an image and converts it into patch embedding, for example, by using a $16\times 16$ 2D convolutional kernel, with a $16\times 16$ stride. This results in a sequence of patches as the image representation, e.g., 196 for a $224\times 224$ input image. Given a video $V\in \mathcal{R}^{T\times H\times W\times C}$, prior approaches either used the same, dense 2D patches (e.g., TimeSFormer \cite{bertasius_arxiv_2021}) or used dense 3D kernels, e.g., 2 or $4\times 16\times 16$ as in ViViT \cite{arnab2021vivit}. In both cases, this results in significantly more tokens, e.g., $T*196$, where $T$ is the number of frames. These tubes or patches are then linearly projected into an embedding space, $z_i\in \mathcal{R}^d$. This sequence of tokens is then processed by a transformer encoder, using standard components, MSA - the multi-head self attention and MLP - the standard transformer projection layer (LN denotes Layer Norm). For a sequence of layers $l\in [0, 1, \ldots L]$, we compute the representation $y_i^l$ and next token features $z_i^l$ for all the $z_i$ tokens:
\begin{equation}
    y_i^l = \text{MSA}(\text{LN}(z_i^{l-1})) + z_i^{l-1}
\end{equation}
\begin{equation}
    z_i^l = \text{MLP}(\text{LN}(y_i^l)) + y_i^l
\end{equation}

To reduce the computational cost, prior approaches factorize the attention mechanism, to have a spatial and temporal attention \cite{arnab2021vivit} or use multiple views with smaller, view level transformers \cite{yan2022multiview}.


\subsection{Sparse Video Tubes}
We propose a simple and straightforward method which is seamlessly applicable to both images and videos. Our approach follows the standard ViT tokenization approach for images: a 2D convolution with a $16\times 16$ kernel. We build on the observation that sparseness is effective for videos. Rather than following the prior works that densely tokenize the video, we instead use the same 2D kernel, but with a large temporal stride, for example, applied to every 16th frame. Thus for an input video clip of $32\times 224\times 224$, this results in only 392 tokens, rather than the ~6k in TimeSFormer or 1-2k in ViViT.

However, this sparse spatial sampling might lose information, especially for quick or short actions. Thus, we create sparse tubes of different shapes, for example, a $16\times 4\times 4$ tube to obtain information from many frames at low spatial resolution.  These tubes can have any shape, and we experimentally explore the effect of these. Importantly, these tubes also have large strides, sparsely sampling the video in different views. We also optionally add an offset to the start location, so that the patches do not always start at $(0,0,0)$ and this allows a reduction in the overlap between the tubes. This is illustrated in Figure \ref{fig:sparse_tubes}. Tubes of various sizes are also used in the MultiView approach for video classification ~\cite{yan2022multiview}, however there they are densely sampled and processed by multiple transformers, resulting in a more computationally intensive approach. 

Furthermore, in contrast to prior works, we also allow for overlap between the tubes. Specifically, we can represent a tube as $(T\times H\times W)$ for the kernel shape, $(T_s, H_s, W_s)$ for the spatio-temporal stride applied to the kernel, and $(x, y, z)$ as the offset of the starting point of the convolution. 

With the proposed design, our approach enables seamless fusion of the image- and video- visual information. The sparse spatial sampling allows sharing the image and frame tokens and the sparse video tubes create a low number of video-specific tokens. This enables better sharing of the ViT model between images and videos.

\subsection{Positional embedding for sparse video tubes}
A key aspect of our approach is the implementation of the positional embedding. In language models, relative positional embeddings are a common and effective approach \cite{vaswani2017attention, what}. However, here, the relative position between two tokens has minimal meaning, and no real reference to where the patch/tube came from in the original video or image. The ViT model \cite{dosovitskiy2020image} and similarly TimeSFormer \cite{bertasius_arxiv_2021} and ViViT \cite{arnab2021vivit} used learnable positional embeddings for the patches. Here, such an approach can be hard for the model, as these learned embeddings do not necessarily reflect where the patches came from in the original video, especially in the case where patches overlap. 

Instead, we use a fixed sine/cosine embedding. Importantly, we take into account the stride, kernel shape and offsets of each tube when applying the positional embeddings. This ensures that the positional embedding of each patch and tube has the global spatio-temporal location of that tube.

Specifically, we compute the embeddings as follows. Here $\tau$ is a constant hyperparameter (we used 10,000). For $j$ from 0 to $d//6$ ($d$ is the number of features), and for $t,x,y$ from 0 to $T, H, W$, $z_i\in \mathcal{R}^{T\times H\times W\times D}$: 
\begin{align}
    \omega_j &= 1 / (\tau^{j}) \\
    p_{j,t} &= \sin (t*\omega_j), \cos (t*\omega_j) \\
    p_{j,x} &= \sin (x*\omega_j), \cos (x*\omega_j) \\
    p_{j,y} &= \sin (y*\omega_j), \cos (y*\omega_j) \\
    z_{i}[t,x,y&,6j:6(j+1)] \mathrel{{+}{=}} [p_{j,t}, p_{j,x}, p_{j,y}]
\end{align}
This adds each spatio-temporal position embedding to the feature dimension of the token $z_i$. Following previous work \cite{vaswani2017attention}, this is done for different wavelengths for each channel. $d//6$ is used since we have 6 elements (a sine and cosine value for each $x,y,t$), this creates a position value for each channel of the representation.

Importantly, here $z_{i}[t,x,y]$ represents the center of the tube, taking into account any strides and offsets used in the tube construction (the channel dimension is not shown here).

After the tokenization step, we concatenate all the tokens together and apply a standard transformer model. This simple structure lets the model share the majority of the weights between all inputs, which we find to be quite beneficial.

\subsection{Sparse Tube Construction}
We explore several methods to create the visual tubes. Our core approach consist of 2 tubes: the $1\times 16\times 16\times d$ tube used to tokenize the image and a $8\times 8\times 8\times d$ tube additionally used for the video. Both have strides of $16\times 16\times 16$. This base tokenizer provides strong performance, but we explore several variations on it.

\textbf{Multi-Tube}. We add multiple tubes to the core approach of various sizes. For example, we can add temporally long and spatially small tubes, such as $16\times 4\times 4$ to learn long actions, or more spatially focused tubes such as a $2\times 16\times 16$ tube. There are many variations of tube shape and stride, which we experimentally explore.

\begin{figure}
    \centering
    \includegraphics[width=0.99\linewidth]{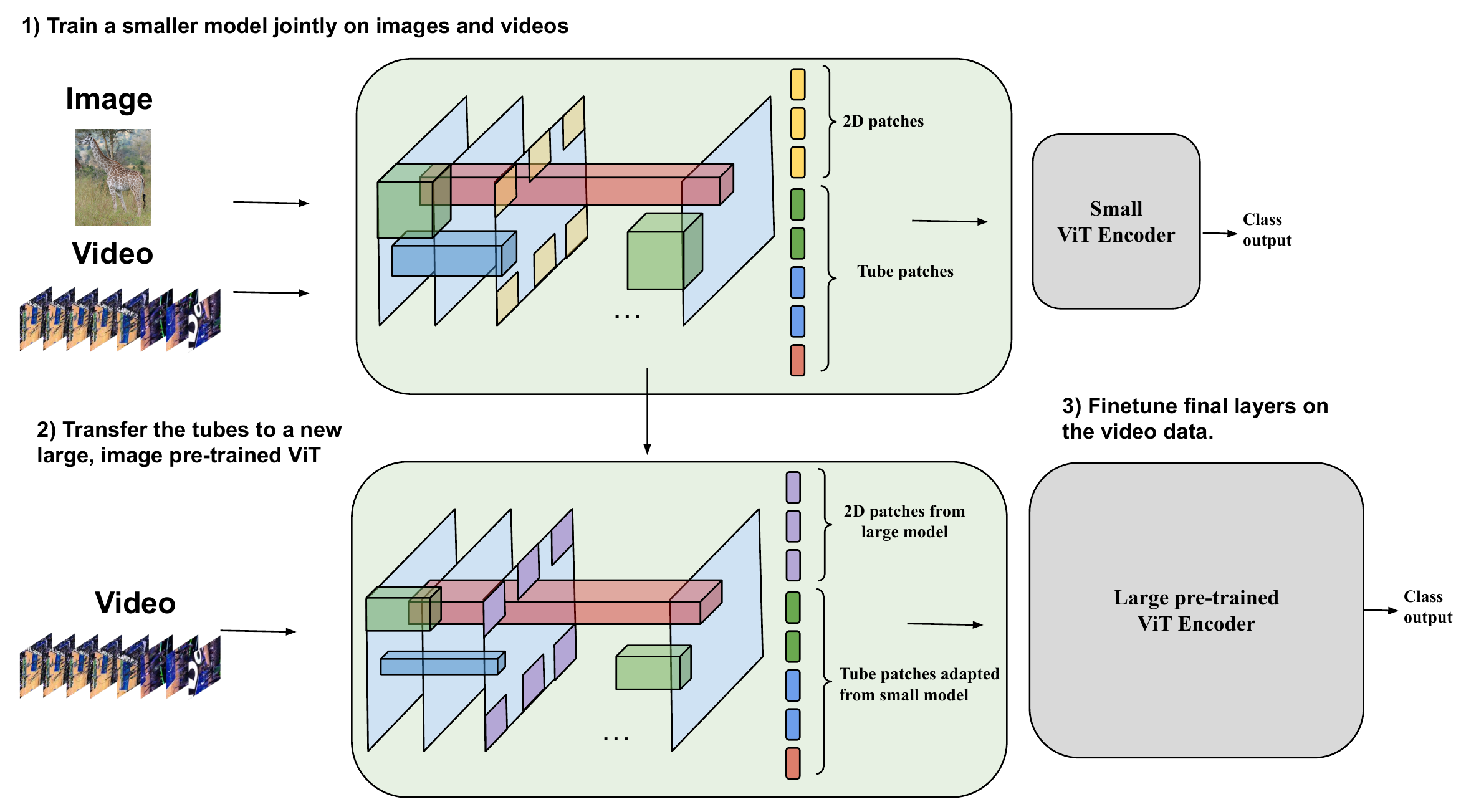}
    \caption{Scaling of TubeViT models: building large scale video models is expensive. We propose to expand model capacity for video models leveraging large pre-trained ViTs.  With TubeViT we can easily train on both image and video data a small-scale model. Then we can adapt the sparse video tubes to a much larger image-only trained ViT, which can be mostly frozen.}
    \label{fig:teaser2}
\end{figure}

\textbf{Space-to-Depth} Another way to extend the core approach is a method inspired by depth-to-space \cite{depth2space}. Here, we reduce the number of channels in a tube, e.g., by a factor of 2. Thus the tube shape becomes $T\times H\times W\times d/2$. Next, we concatenate 2 tokens along the channel axis. We can then also reduce the stride of the tube. This results in the same number of tokens and dimensions as the original, but effectively increases the kernel size without changing the number of parameters. I.e., when the stride is reduced on the time axis, the token now represents $T*2\times H\times W$ locations, but only uses $T*H*W$ parameters. In the experiments, we explore different settings: e.g., more temporal dense vs more spatially dense and the depth to space factor (2, 4, 8, etc.). 

\textbf{Interpolated Kernels}. For this setting, rather than having a unique kernel for each tube, we learn 1 3D kernel of shape $8\times 8\times 8$. We then use tri-linear interpolation to reshape the kernel to various sizes, e.g., 4x16x16 or 32x4x4, etc. depending on the tube configuration. Any sized kernel can be created from this single kernel. This method has several advantages. (1) It reduces the number of learned parameters that are only used on the video stream. (2) It enables more flexible usage of the kernels, e.g., it can be made longer to handle longer videos, or spatially larger to find small objects.


The TubeViT approach consists of the union of the above-mentioned Multi-Tube and Space-to-Depth, the exact settings are provided in the supplemental materials. We experiment with Interpolated Kernels in ablations.

\subsection{Image and Video Joint Training} As described above, our approach  seamlessly adapts to either image, video or both inputs. While image+video joint inputs are rare, the ability to use them together while training is very important as many datasets with valuable annotations (e.g., ImageNet, Kinetics) come from either image sources or video sources but not both. Jointly training with our approach is easy -- the image is tokenized by the 2D kernel and the video is tokenized by both the 2D patches (with large temporal stride) and Sparse Tubes. Both are then passed into a standard ViT; the position embedding will be supplied in either case.  The position embedding approach is also needed for the joint training to be effective. We demonstrate the benefits of our approach for joint training in the experiments, Section~\ref{sec:experiments}.


\subsection{Image-To-Video Scaling Up of Models}
\label{sec:scaling}
We also propose a method for a more efficient way of scaling up the models (Figure~\ref{fig:teaser2}). Training large ViT models is computationally expensive, especially for videos. Since nearly all the components of our model are shared between the both images and videos, we explore a method to utilize large models without having heavy fine-tuning.

First, we train a smaller model jointly on images and videos. This gives us a set of weights for the tubes. Then we take a large pre-trained image ViT, but further add the tubes. These tubes use the same kernel weights as the smaller model, and so we can avoid further training them. Since larger ViTs generally use more channel dimensions than smaller ones, we use the space-to-depth transform again here to create tokens with the proper channel dimensions without needing new weights.

Next, we pick a point in the network and freeze all the layers before it, for example, the 26th of 32 layers in ViT-H. At this point, we add a gated connection to the network:
\begin{equation}
\label{eq:gate_scale}
    z^{s} = \text{MLP}(\text{LN}(y^{s})) + y^{s} + \tanh(\alpha)z^0
\end{equation}
where $s$ is the layer the network is frozen at (e.g., 26) of the ViT model and $z^0$ is the raw input tokens from the tubes. $\alpha$ is the learned gating parameter, initialized at 0. In the first steps of training, this gate has no effect on the representation, and thus the ViT is unchanged. However, it can learn to incorporate the raw tubes at this point and further refine the later weights.  


\begin{table}
		\centering
    	\setlength{\tabcolsep}{3pt} %
		\renewcommand*{\arraystretch}{1.10}  %
		
		\vspace{-0.3\baselineskip}
		\scriptsize{
			\begin{tabular}{lccccc}
				\toprule
				Method   & PT Data  & Top 1                & Top 5         & Crops 	& TFLOPs \\
				\midrule
				TSM-ResNeXt-101~\cite{lin_tsm_cvpr_2019}			& ImageNet-1k			  & 76.3				 & -- &  -- & -- \\%
				I3D NL~\cite{wang2018nonlocal}	& ImageNet-1k									 & 77.7                  & 93.3         		 &  $10 \times 3$ & 10.77     \\ 
				VidTR-L~\cite{zhang2021vidtr} & ImageNet-1k & 79.1 & 93.9 & $10 \times 3$ & 10.53 \\
				LGD-3D R101~\cite{qiu2019learning} & ImageNet-lk							&  79.4				    & 94.4			 		&  --	& --					\\  
				SlowFast R101-NL~\cite{feichtenhofer_iccv_2019} & -       		&  79.8                 &  93.9                   & $10 \times 3$    & 7.02   \\  %
				X3D-XXL~\cite{feichtenhofer_cvpr_2020} & -      					&  80.4					&  94.6			  		& $10 \times 3$   & 5.82   \\
				OmniSource~\cite{duan2020omni} & ImageNet-1k & 80.5 & 94.4 & -- & -- \\
				TimeSformer-L~\cite{bertasius_arxiv_2021} & ImageNet-21k & 80.7				& 94.7					& $1 \times 3$ & 7.14 \\
				MFormer-HR~\cite{patrick2021keeping} & ImageNet-21k & 81.1 & 95.2 & $10 \times 3$ & 28.76 \\
				MViT-B~\cite{fan2021multiscale} & -					  	& 81.2				& 95.1					& $3 \times 3$ & 4.10 \\
				MoViNet-A6~\cite{kondratyuk2021movinets} & - & 81.5 	&  \textbf{95.3}  & $1 \times 1$ & 0.39  \\  %
				ViViT-L FE~\cite{arnab2021vivit} & ImageNet-1k & 81.7 	&  93.8  & $1 \times 3$ &  11.94 \\  %
				MTV-B \cite{yan2022multiview} & ImageNet-21K & 82.4  & 95.2 	& $4 \times 3$ & 11.16 \\ %
				VideoMAE  \cite{tong2022videomae} & - & 87.4	& 97.6 & - & -\\
				\midrule
				\multicolumn{4}{l}{\textit{Large Scale Pretraining Data}}                                \\ 
				VATT-L~\cite{akbari2021vatt} & HowTo100M  &  82.1 &  95.5 	& $4 \times 3$ & 29.80 \\  %
				ip-CSN-152~\cite{tran_iccv_2019} & IG-65M &  82.5					& 95.3			&  $10 \times 3$ & 3.27	  \\
				R3D-RS~\cite{du2021revisiting} & WTS & 83.5 & -- & $10 \times 3$ & 9.21  \\
				OmniSource~\cite{duan2020omni} & IG-65M & 83.6 & 96.0 & -- & -- \\
				MAE-ST \cite{feichtenhofer2022masked} & IG-1M & 84.4 & - & - & - \\
				ViViT-H~\cite{arnab2021vivit} & JFT														&  84.9 &  95.8 	& $4 \times 3$ & 47.77 \\  %
				TokenLearner-L/10~\cite{ryoo2021tokenlearner_neurips} & JFT														&  85.4 &  96.3 	& $4 \times 3$ & 48.91 \\  %
				Florence~\cite{yuan2021florence} & FLD-900M & 86.5 & 97.3 & $4 \times 3$ & -- \\
				CoVeR ~\cite{zhang2021co} & JFT-3B & 87.2 & -- & $1 \times 3$ & -- \\
				CoCa \cite{coca} & ALIGN (1.8B) & 	88.9 & - & - & - \\
				MTV-H \cite{yan2022multiview} & WTS 280p														&  89.9 & 98.3 	& $4 \times 3$ & 73.57 \\ %
				\midrule
				TubeVit-B & ImageNet-1k & 88.6 & 97.6 & $4\times 3$ & 0.87 \\
				TubeVit-L & ImageNet-1k & \textbf{90.2} & \textbf{98.6} & $4\times 3$ & 9.53 \\
				\midrule
				TubeViT-H (created) & ImageNet-1k & \textbf{90.9} & \textbf{98.9} & $4\times 3$ & 17.64 \\
				\bottomrule
			\end{tabular}
		}
		\caption{Performance on Kinetics 400. TubeViT performs best.  We report the crops and total TFLOPs used for inference. The crops, $t \times x$ denotes $t$ temporal and $x$ spatial crops.}
		\label{tab:sota_kinetics400}
	\end{table}

\begin{table}
		\centering
  		\setlength{\tabcolsep}{4pt} %
		\vspace{-0.3\baselineskip}
	  		\begin{tabular}{lcc}
	  			\toprule
	  			Method 																			 & Top 1                & Top 5      \\ %
	  			\midrule
	  			SlowFast R101-NL~\cite{feichtenhofer_iccv_2019}       		&  81.8                     &  95.1   \\ %
	  			X3D-XL~\cite{feichtenhofer_cvpr_2020}      						 &  81.9					&  95.5		\\ %
	  			TimeSformer-L~\cite{bertasius_arxiv_2021}				  & 82.2				& 95.6		\\ %
	  			MFormer-HR~\cite{patrick2021keeping} & 82.7 & 96.1  \\
	  			ViViT-L FE~\cite{arnab2021vivit} & 82.9 	&  94.6 \\  %
	  			MViT-B~\cite{fan2021multiscale}					  & 83.8				& 96.3		\\ %
	  			MoViNet-A6~\cite{kondratyuk2021movinets} & 84.8 	&  96.5 \\
	  			\midrule
  				R3D-RS~\cite{du2021revisiting} (WTS) & 84.3 & --  \\
	  			ViViT-H~\cite{arnab2021vivit} (JFT) & 85.8 & 96.5 \\ %
	  			TokenLearner-L/10~\cite{ryoo2021tokenlearner_neurips} (JFT) &  86.3 &  97.0 \\  %
	  			Florence~\cite{yuan2021florence} (FLD-900M) & 87.8 & 97.8 \\
	  			CoVeR~\cite{zhang2021co} (JFT-3B) & 87.9 & -- \\
	  			MTV-H \cite{yan2022multiview} (WTS 280p) &  90.3 &  98.5 \\ %
	  			CoCa \cite{coca} (ALIGN 1.8B) & 89.4 & - \\
	  			Merlot-Reserve-L \cite{merlotreserve} (YT-1B) & 91.1 & 97.1 \\
	  			\midrule
				TubeVit-B (ImageNet-1k) & 90.9 & 97.3 \\
				TubeVit-L (ImageNet-1k) & \textbf{91.5} & \textbf{98.7} \\
				\midrule`
				TubeVit-H (created) & \textbf{91.8} & \textbf{98.9} \\
	  			\bottomrule
	  		\end{tabular}
	  		\caption{Performance on Kinetics 600. Similarly, to Table~\ref{tab:sota_kinetics400} our model uses the ImageNet-1k dataset. Most models use significantly larger pre-training datasets (bottom half). Tube-ViT outperforms prior work.}

	  		\label{tab:sota_kinetics600}
	  	\end{table}

\begin{table}
  		\setlength{\tabcolsep}{4pt} %
  		\centering
  		\vspace{-0.3\baselineskip}
  		\setlength{\tabcolsep}{6pt} %
  		\scriptsize{
  			\begin{tabular}{lcc}
  				\toprule
  				& Top 1 & Top 5 \\ 
  				\midrule
  				VidTR-L~\cite{zhang2021vidtr} & 70.2 & -- \\
  				SlowFast R101~\cite{feichtenhofer_iccv_2019} 	&  71.0 & 89.6       \\
  				MoViNet-A6~\cite{kondratyuk2021movinets} 	&  72.3     &  --     \\
  				\midrule
	  			CoVeR (JFT-3B)~\cite{zhang2021co} & 79.8 & -- \\
                CoCa (Align 1.8B) \cite{coca} & 	82.7 & - \\
                MTV-H (WTS 280p) \cite{yan2022multiview} & 83.4 & 96.2 \\  %
                \midrule
                TubeViT-L & \textbf{83.8} & \textbf{96.6} \\
  				\bottomrule
  			\end{tabular}
      		\caption{Performance compared to SOTA on Kinetics 700.}
  			\label{tab:sota_kinetics700}
  		}
\end{table}

\begin{table}
  		\setlength{\tabcolsep}{4pt} %
  		\centering
  		\vspace{-0.3\baselineskip}
  		\setlength{\tabcolsep}{6pt} %
  		\scriptsize{
  			\begin{tabular}{lcc}
  				\toprule
  				& Top 1 & Top 5 \\ 
  				\midrule
  				SlowFast R50~\cite{feichtenhofer_iccv_2019} 	&  61.7 & --       \\
  				TimeSformer-L~\cite{bertasius_arxiv_2021}  & 62.5	\\
  				VidTR-L~\cite{zhang2021vidtr} & 63.0 & -- \\
  			    CoVeR~\cite{zhang2021co} & 64.7 & -- \\
  			    MoViNet-A3~\cite{kondratyuk2021movinets} 	& 64.1 &88.8  \\
  			    ViViT-L FE~\cite{arnab2021vivit} & 65.9 	&  89.9 \\  %
	  		
	         	VoV3D-L~\cite{lee2020diverse}					  &  67.3 & 90.5		\\
	  		    MFormer-L~\cite{patrick2021keeping}					  &  68.1 & 91.2		\\
  		        MTV-B (320p) \cite{yan2022multiview} &  68.5 & 90.4 \\  %
                MViT-B~\cite{fan2021multiscale}					  & 68.7				& 91.5		\\
                MViT \cite{li2022mvitv2} & 73.3	& 94.1 \\ 
                MaskFeat \cite{wei2021maskedfeature} & 75.0 & 95.0 \\
                VideoMAE \cite{tong2022videomae} & 75.4 & 95.2 \\
                \midrule
                TubeViT-L & \textbf{76.1} & \textbf{95.2} \\
  				\bottomrule
  			\end{tabular}
      		\caption{Performance on Something-SomethingV2 dataset.}
  			\label{tab:something}
  		}
\end{table}



\section{Experiments}
\label{sec:experiments}
We evaluate the approach on several popular datasets: Kinetics 400, Kinetics 600, Kinetics 700 \cite{kay_arxiv_2017,carreira2019short}, and SomethingSomething V2 \cite{somethingsomething}. 
These datasets cover a wide variety of video understanding challenges and are well established in the literature. The main results are trained jointly on ImageNet-1k (of 1.2M images) and the video data, please see the supplemental materials for full details. We use standard Top 1 and Top 5 evaluation metrics and report FLOPs of ours and previous works, when available. 
Our model sizes are \textbf{90M Base (B)}, \textbf{311M Large (L)}. A \textbf{635M Huge (H)} is `created' with Image-to-Video scaling.


\subsection{Main results}
For the main results, we use 4 tubes with the following configuration (order of $t, h, w$): (1) $8\times 8\times 8$ with a stride of $(16, 32, 32)$; (2) $16\times 4\times 4$ with a stride of $6\times 32\times 32$ and an offset of $(4, 8, 8)$; (3) $4\times 12\times 12$ with a stride of $(16, 32, 32)$ and an offset of $(0, 16, 16)$; and (4) $1\times 16\times 16$ with a stride of $(32, 16, 16)$. For an input of $32\times 224\times 224$, this results in only 559 tokens, significantly less than other approaches. In the supplemental material, we have detailed experiments over many tube configurations, as well as the space-to-depth settings used.

We would like to note that with data augmentation such as random spatial and temporal cropping, over multiple training epochs the model will see different parts of the video, even with sparse sampling.

\textbf{Comparison to SOTA.}
First, we compare our final approach to previous state-of-the-art (SOTA) methods. Tables~\ref{tab:sota_kinetics400},~\ref{tab:sota_kinetics600} and \ref{tab:sota_kinetics700} shows the performance of our model compared to the state-of-the-art on the Kinetics-400 Kinetics-600 and Kinetics-700 datasets.
Table~\ref{tab:sota_kinetics400} shows additional information (e.g. views, pre-training datasets) which applies to the other tables as well. These results show our approach outperforms SOTA, both in terms of accuracy and efficiency. We also outperform methods on co-training of images and videos, and methods with strong video pre-training. 

We note that all the sizes of our model perform well, despite the fact that others are much larger or use significantly larger pre-training data (e.g., CoCa with 1B params and 1.8B examples, MerlotReserve has 644M params and uses YT-1B dataset). 
Table~\ref{tab:something} shows our results on the Something-Something dataset (SSv2). This dataset is often used to evaluate more dynamic activities. Our approach outperforms SOTA on it as well.


\textbf{Joint image+video training.}
We further explore the effects of co-training on image+video datasets, finding this to be highly effective as also shown above.  Table~\ref{tab:im_kin} evaluates this in a side-by-side experiment of using Kinetics (video) only vs Kinetics and ImageNet datasets for pre-training. We see that there is a large gain from the co-training of our approach. 
We see that two-stage training, i.e., first training on one dataset and then training on a second one, is also weaker than the joint training, as the two datasets cannot interact during training.
We also compare to prior methods such as TimeSFormer \cite{bertasius_arxiv_2021} only using dense 2D patches, or using inflated 3D kernels (e.g., ViViT \cite{arnab2021vivit}). In both cases, we see a clear benefit from the proposed approach. We also note that these prior approaches have significantly more FLOPs, due to the large number of tokens from the dense sampling. Our observations that image and video co-training is beneficial are consistent with prior works \cite{zhang2021co,likhosherstov2021polyvit}; here the difference is that we have a single compact model to do that. 

As a sanity check, we also compare our performance on ImageNet-1k, without any hyperparameter tuning or additions: our ViT-B model only trained on ImageNet has 78.1 accuracy, similar to the ViT-B in \cite{steiner2021augreg}. When joint training with Kinetics-600, the model gets 81.4, a gain of 3.4\%, showing the benefits of joint training for image-only tasks too. While other works achieve higher performance on ImageNet, they often use specialized data augmentation, learning schedules, and other tricks which we are not using. Instead, we are purely studying the benefit from using both videos and images.

\begin{table}[]
    \centering
    \begin{tabular}{l|c}
         & Kinetics 600 \\
    \midrule
    TubeViT-L Kinetics-only     & 85.6 \\
    TubeViT-L ImageNet then Kinetics & 90.4 \\
    TubeViT-L ImageNet+Kinetics Jointly & \textbf{91.5} \\
   \midrule
   2D Patches only ImageNet+Kinetics & 87.6 \\
   Inflated 3D Patches ImageNet then Kinetics & 88.4 \\
    \midrule
    \end{tabular}
    \caption{Combining datasets, which TubeViT seamlessly allows, is highly effective, as seen here in these side-by-side results for the Kinetics-600 dataset. The results are based on the ViT-L model.}
    \label{tab:im_kin}
\end{table}

\textbf{Scaling video training with sparse video tubes.}
In Table \ref{tab:new_scaling} we demonstrate how a small TubeViT model can be adapted leveraging a large and (often independently) pre-trained model on images only. We start by leveraging a large, image-pretrained ViT, here ViT-H. We then take the learned tubes from TubeViT-B and use them along with the ViT-H image tokenizer to generate a set of tokens from a video, same as before. Then these are used as input to ViT-H, and we finetune only the latter parts of the model on the video data. 
These results suggests that this is an effective way to scale and utilize giant ViT models without needing the high compute cost to fully finetune the model. We also see that the gating in Eq. \ref{eq:gate_scale} is effective. We also found that in this setting, training time was reduced by 43\%, as it has fewer weights to update.

\begin{table}[]
    \centering
    \begin{tabular}{l|c}
    Models & K600, Accuracy (\%) \\
    \midrule
     TubeViT-H Full Finetune & 91.8 \\
     \midrule
     \multicolumn{2}{l}{Scaling method with different portions trained}\\
     Last FC Layer & 85.6 \\
     + Last 4 Layers & 86.3 \\
     + Last 8 Layers & 86.8 \\
     + Last 8 + Gated (Eq. \ref{eq:gate_scale}) & 89.7 \\
    \bottomrule
    \end{tabular}
    \caption{Image-to-Video Scaling. We take a ImageNet pre-trained ViT-H and use a set of Tubes from TubeViT-B to create the tokens. We then fine-tune different portions of the model to see how we can best take advantage of existing, large pretrained ViT models. Even pretraining of handful of layers can achieve performance approaching the full model training.}
    \label{tab:new_scaling}
\end{table}

\begin{figure}
    \centering
    \includegraphics[width=\linewidth]{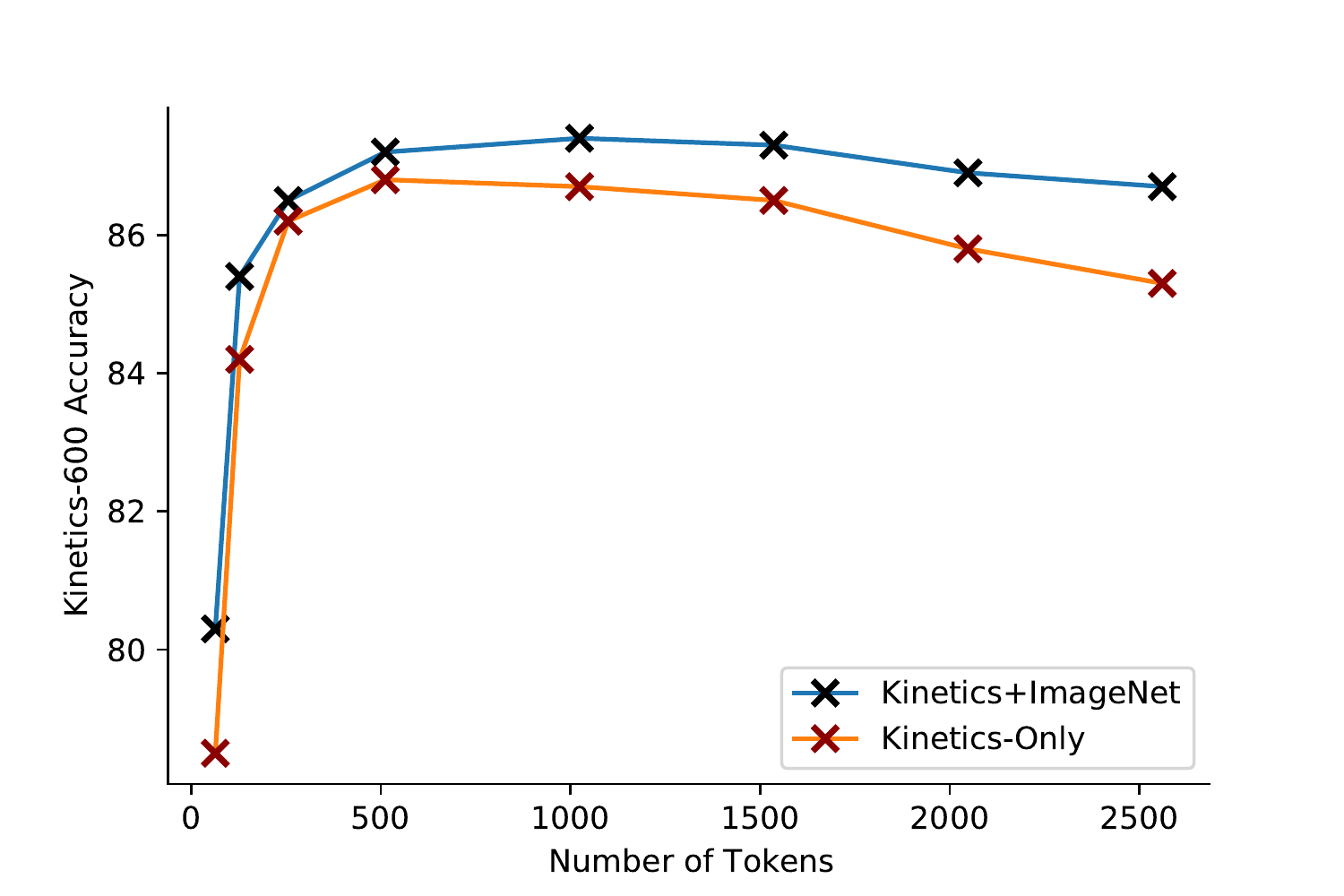}
    \caption{Accuracy vs. Number of tokens used in our model. We find that when increasing the tokens above 1500, there is a noticeable drop in performance, especially when only training on Kinetics-600 data. Joint training is more robust.}
    \label{fig:num_tokens}
\end{figure}

\textbf{Detrimental Effects of Too Many Tokens.} Next we study the effect of number of tokens used in the model, shown in Figure \ref{fig:num_tokens}. This result is another key insight as to why our approach works so well: with too many tokens, the performance drops, especially when only using Kinetics data. There are a number of possible reasons for why this occurs, for example, the self-attention mechanism could be struggling to learn for longer sequences, or there may not be sufficient data to learn the longer sequences, or perhaps the model is overfitting with longer sequences. This result indicates that for current datasets, the sparse sampling is an effective and efficient way to process videos. Further, it is possible that existing using long, densely sampled sequences are effected by this, and perhaps another reason the factorized attention modules are needed.

\subsection{Ablations}
\label{sec:ablations}
In this section, we present a number of ablation studies to determine why this method is effective. For these experiments we use Kinetics 600.

\textbf{Main ablations.}
First, we study the effect of the choice of position biases (Table \ref{tab:ablation:pos_emb}). We find that adding fixed cosine position embedding performs best and much better than other embeddings. Intuitively, this makes sense, since we are sparsely sampling potentially overlapping tokens, this method is able to best capture the token location.

Next in Table \ref{tab:ablation:num_tubes}, we study the number of tubes used. This finding, which is consistent with previous multi-view observations \cite{yan2022multiview}, shows that having a variety of tubes is beneficial to video understanding. 

Next, in Table \ref{tab:ablation:d2s}, we study the depth-to-space versions of the network. Here, we reduce the channels of the generated tokens from $D//S$, e.g., by a factor of 2 or 4. Then after generating the tokens, we concatenate them along the channel axis. We study both increasing the number of tokens along the spatial and temporal dimensions. We find this to be an effective method, as it enables more dense samples without increasing the number of parameters or tokens.


Table \ref{tab:ablation:evaltokens} compares evaluating with more patches than the model was trained with. To do this we reduce the strides of the kernel. Initially this improves results, but after increasing 2x, the performance begins to drop, likely because the evaluation data is too different from the training one.

In Table \ref{tab:ablation:interpolate}, we study the ability of the interpolated single kernel. I.e., rather than having $N$ 3D convolutional kernels, one for each tube, we build 1 $8\times 8\times 8$ 3D kernel and use interpolation to generate the different tube shapes. Somewhat surprisingly, we find this works fairly well, while also reducing the number of learnable parameters in the network.

In Table \ref{tab:ablation:eval_views}, we compare the approach with different number of temporal and spatial crops. We find that even a single crop gives strong performance, and the standard $4\times 3$ performs nearly the same as the $10\times 10$ setting, suggesting that the sparse samples are quite suitable and further information is not as beneficial.

\begin{table*}[]
\centering 
\begingroup
		\captionsetup[subfloat]{width=.25\linewidth,captionskip=5pt}
		\subfloat[\textbf{Position Embeddings}. Fixed, cosine embeddings with strides is best.
		\label{tab:ablation:pos_emb}]{\tablestyle{2pt}{1.05}
		\begin{tabular}{l|c}
		& K600  \\
		\midrule
		None &  78.6 \\
		Learned &   79.2 \\
		Relative &   77.5 \\
		Fixed Cosine (no stride) &   77.7 \\
		Fixed Cosine (Ours) &  84.5 \\
		\bottomrule
		\end{tabular}}\hspace{15mm}
		\endgroup
		\begingroup
		\captionsetup[subfloat]{width=.15\linewidth,captionskip=5pt}
			\subfloat[\textbf{Number of Tubes}. 
		\label{tab:ablation:num_tubes}]{\tablestyle{2pt}{1.05}\fontsize{9}{10.2}\selectfont
		\begin{tabular}{l|c|c}
		& GF & K600  \\
		\midrule
		1 & 70 & 78.4 \\
		2 & 71 & 81.5 \\
		4 & 72 & 83.4 \\
		8 & 74 & 85.4 \\
		\bottomrule
		\end{tabular}}\hspace{1.8cm}
		\endgroup
		\begingroup
		\captionsetup[subfloat]{width=.25\linewidth,captionskip=3pt}
			\subfloat[\textbf{Space To Depth}. Applying space-to-depth temporally (T), spatially (S), and spatio-temporally (ST).
		\label{tab:ablation:d2s}]{\tablestyle{2pt}{1.05}
		\begin{tabular}{l|c|c}
		& GF & K600  \\
		\midrule
		Baseline & 72 & 83.4 \\
		With D2S x2 T & 72 & 84.7 \\
		With D2S x2 S & 72 & 84.5 \\
		With D2S x4 T & 72 & 85.1 \\
		With D2S x4 S & 72 & 85.4 \\
		With D2S x4 ST & 72 & 85.3 \\
		\bottomrule
		\end{tabular}}\vspace{-3mm}\\ \hspace{1cm}
		\endgroup
		\begingroup
		\captionsetup[subfloat]{width=.2\linewidth,captionskip=3pt}
        \subfloat[\textbf{Eval Tokens}. 
			Generating larger number of tokens at eval time than in training, where 559 are used.
		\label{tab:ablation:evaltokens}]{\tablestyle{2pt}{1.05}\fontsize{8}{9.2}\selectfont
		\hspace{6mm}
		\begin{tabular}{l|c}
		&  K600  \\
		\midrule
		Base (559)  & 84.5 \\
		768 & 84.9 \\
		1024 & 84.6 \\
		1536 & 83.5 \\
		\bottomrule
		\end{tabular}}\hspace{2.5cm}
		\endgroup
		\begingroup
		\captionsetup[subfloat]{width=.18\linewidth,captionskip=3pt}
			\subfloat[\textbf{Interpolated Kernel}. Using a single 3D kernel interpolated to different sizes.
		\label{tab:ablation:interpolate}]{\tablestyle{2pt}{1.05}
		\begin{tabular}{l|c}
		&  K600  \\
		\midrule
		Interpolated & 83.8 \\
		TubeViT & 84.5 \\
		\bottomrule
		\end{tabular}}\hspace{2cm}
		\endgroup
		\begingroup
		\captionsetup[subfloat]{width=.2\linewidth,captionskip=3pt}
		\subfloat[\textbf{Multi-Crop Evaluation}. $4\times 3$ is used in the paper.
		\label{tab:ablation:eval_views}]{\tablestyle{2pt}{1.05}\hspace{10mm}
		\begin{tabular}{l|c}
		& K600  \\
		\midrule
		$1\times 1$ & 82.8 \\
		$4\times 1$ & 83.3 \\
		$1\times 3$ & 83.6 \\
		$4\times 3$ & 84.5 \\
		$10\times 10$ & 84.7 \\
		\bottomrule
		\end{tabular}}
		\endgroup
\caption{Ablation studies on various components of our approach on Kinetics-600, using TubeViT-B.}
\label{tab:ablations}
\end{table*}

\textbf{Factorized attention ablations.}
In Table \ref{tab:added_layers}, we further study the effect of adding a new attention layer to an ImageNet pre-trained ViT model. Here, we are using the tube method to tokenize the inputs, but instead of using a factorized attention module, we simply add an additional self-attention layer. This has a similar effect of the factorized attention approaches that add new, uninitialized $K,Q,V$ projections to a pre-trained ViT (e.g., TimeSFormer and ViViT). These results indicate that such methods are not able to best utilize the image pre-trained weights of the network due to these new layers. Since the sparse tubes yield few additional tokens, they can directly use the same ViT model without factorized attention and are thus able to better utilize the image trained weights. Note that there are still differences between the works, e.g., the reduced number of tokens, etc. However, we believe this observation holds, and is a possible explanation for why the spatio-temporal attention in ViVit performed better for some datasets.


\begin{table}[]
    \centering
    \vspace{-0.1cm}
    \begin{tabular}{l|c}
    Layers Added & K600 \\
    \midrule
    0     & 84.23 \\
    \midrule
    1 & 80.23 \\
    2 & 78.87 \\
    4 & 75.24 \\
    8 & 72.95 \\
    \bottomrule
    \end{tabular}
    \caption{We find that adding even a single layer to a pretrained image network degrades performance. This suggests that the factorized attention methods are sub-optimal since they cannot fully take advantage of the image-pre-trained networks. Trained for 70k steps.}
    \label{tab:added_layers}
\end{table}

 \textbf{Model scaling ablations.}
 Table~\ref{tab:pt_scaling} provides ablations on scaling to create TubeViT Base from a Tiny one. Even just training the final few layers is effective (4 of 12), and can nearly match the performance of full finetuning. 
 This is consistent with our observations in Table~\ref{tab:new_scaling} for ViT-H.

 \begin{table}[]
     \centering
     \begin{tabular}{l|c}
     Trained & K600 \\
     \midrule
     Last FC Layer & 79.6 \\
     + 1 Layer & 80.8 \\
     + 4 Layers & 81.1 \\
     Whole Model     & 81.4 \\
     \bottomrule
     \end{tabular}
     \caption{Image-to-Video scaling from Tiny to Base. We take a ImageNet pre-trained ViT-Base and the TubeViT corresponding to ViT-Tiny and ImageNet pre-trained ViT-Base to create a larger TubeViT. 
     These models were trained for 50k steps.}
     \label{tab:pt_scaling}
 \end{table}

\begin{figure}
    \centering
    \includegraphics[width=\linewidth]{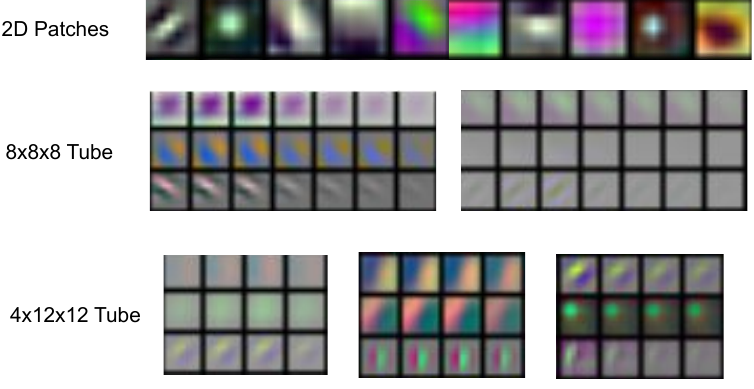}
    \caption{Visualization of a selected set of 2D patches and tubes.}
    \label{fig:learned_tube_vis}
\end{figure}

Figure~\ref{fig:learned_tube_vis} visualizes the learned 2D patches and 3D tubes. 

\section{Conclusion}
We proposed sparse video tubes for video recognition. With sparse video tubes, a ViT encoder can be transformed into an efficient video model. The approach is simple, enables seamless joint training with images and videos and improves video recognition across multiple datasets. We also demonstrate an elegant scaling of video models with our proposed method. We conduct extensive ablation experiments to determine why the approach works, finding the a combination of the joint training, reduced tokens, and better utilization of shared image+video weights led to the improvements. We obtain SOTA or above performance.

\clearpage
\newpage
\appendix

\section{Implementation Details}
Our hyperparameters are summarized in Table \ref{tab:training_hyperparameters}. For all datasets, we employ random spatial and temporal cropping. For most datasets, these settings were the same. For Charades, we decreased the batch size but used longer, 128 frame clips, as Charades videos are roughly 30 seconds long, compared to 10 seconds for Kinetics. 

We also found some training instability when using larger ViT models. When using ViT-L or ViT-H models, we had to decrease the weight decay value as well as the learning rate, otherwise we found the training accuracy dropped to 0 and the loss stayed flat.

For smaller datasets, such as Charades and SSv2, we had to increase the data augmentation settings, as done in previous works, e.g., \cite{yan2022multiview}. We added Mixup and label smoothing and dropout to them.

For all the datasets, we applied RandAugment \cite{cubuk_arxiv_2019}, as we found this to be beneficial. We also kept the number of steps the same for all datasets.

\begin{table*}[t]
\centering
\begin{tabular}{lccccc} %
\toprule
                              & K400 & K600 & K700 & Charades & SSv2 \\ \midrule
\multicolumn{6}{l}{\textit{Optimization}}                                                                                 \\
Optimizer               & \multicolumn{5}{c}{Adam}                                                               \\
Batch size & 256 & 256 & 256 & 64 & 256 \\
Learning rate schedule	& \multicolumn{5}{c}{cosine decay + linear warmup} \\
Linear warmup steps	& \multicolumn{5}{c}{10,000} \\
Base learning rate			& \multicolumn{3}{c}{5e-5 (L,H 1e-5)} & 1e-3 & 2e-5 \\
Steps	& \multicolumn{5}{c}{300,000} \\
\midrule
\multicolumn{6}{l}{\textit{Data augmentation}} \\
Rand augment number of layers~\cite{cubuk_arxiv_2019}		 & \multicolumn{5}{c}{2} \\
Rand augment magnitude~\cite{cubuk_arxiv_2019} & \multicolumn{5}{c}{10}  \\
Weight Decay & \multicolumn{5}{c}{0.001 (B) 1e-5 (L,H)}\\
Mixup \cite{zhang_mixup_iclr_2018} & - & - & - & - & 0.3 \\
Dropout & - & - & - & 0.2 & 0.3 \\
Label Smoothing & - & - & - & 0.1 & 0.3 \\
\midrule
Number of Frames & 64 & 64 & 64 & 128 & 32 \\
FPS  & 15 & 15 & 15 & 6 & 24 \\
\bottomrule
\end{tabular}
\caption{Training hyperparamters for our experiments. We note when different settings were used for the base (B), large (L) and huge (H) models.}
\label{tab:training_hyperparameters}
\end{table*}

\textbf{Joint ImageNet and Kinetics Training.}
When jointly training on the two (or more) datasets, we use a separate fully connected layer to output the class predictions. E.g., for ImageNet and Kinetics-600, we use an FC layer with 1,000 and 600 outputs. We then compute the loss for the relevant head and backpropagate it. During the joint training, we use the same settings as listed in Table \ref{tab:training_hyperparameters}. We use the joint training for Kinetics 400, 600 and 700. For Charades and SSv2, we use the Kinetics-600+ImageNet pretrained model and finetune it on the dataset.

\textbf{Full Model Settings.}
Our model is based on the standard ViT models, thus the core of the approach is the same as previous ViTs \cite{dosovitskiy2020image}. We summarize those settings in Table \ref{tab:vits}.

\begin{table*}[]
    \centering
    \begin{tabular}{cccccc}
    \toprule
    Model & Layers & Hidden size $d$ & MLP size & Num Heads & Params\\
    \midrule
    ViT-Base & 12 & 768 & 3072 & 12 & 86M\\
    ViT-Large & 24 & 1024 & 4096 & 16 & 307M\\
    ViT-Huge & 32 & 1280 & 5120 & 16 & 632M\\
    \bottomrule
    \end{tabular}
    \caption{Parameter count for the vit encoder backbones.}
    \label{tab:vits}
\end{table*}

In Table \ref{tab:tube_Config}, we detail the settings for each tube.

\begin{table*}[]
    \centering
    \begin{tabular}{ccccc}
    \toprule
    Kernel & Stride & Offset & S2D & params \\
    \midrule
    $8\times 8\times 8$  & $(16, 32, 32)$ & (0, 0, 0) & 2x temporal & $512d$ \\
    $16\times 4\times 4$ &  $(6, 32, 32)$ & (4, 8, 8) & 4x spatial & $256d$\\
    $4\times 12\times 12$ & $(16, 32, 32)$ & (0, 16, 16) & - & $576d$\\
    $1\times 16\times 16$ & $(32, 16, 16)$ & (0, 0, 0) & - & $256d$\\
    \bottomrule
    \end{tabular}
    \caption{Configuration for the tubes using the in main Tube-ViT. We also report the number of params used by each tube, which depends on $d$, the hidden size of the ViT model used. The tubes add an additional 1-3M params, depending on the model, a small fraction of the total model size.}
    \label{tab:tube_Config}
\end{table*}

\section{Additional Experiments on Charades}
We include results on Charades \cite{sigurdsson2016hollywood} to show the effectiveness of this approach on longer videos, since Charades videos are on average 30 seconds long. However, Charades is also a multi-label dataset, and we found it required different settings to effectively train, so we include all those details here.

First, we found that the core multi-tube approach were not performing as well as some prior work (e.g., AssembleNet \cite{ryoo2020assemblenet++}). Since Charades has a lot of object-related actions and contains longer videos with more temporal information, we modified the core model to make it more suitable for this data. First, we used the interpolation method to increase the tube shapes to:

\begin{itemize}
    \item $1\times 16\times 16$
    \item  $16\times 16\times 16$
    \item $32\times 8\times 8$
    \item $4\times 32\times 32$
\end{itemize}

We note two important factors. First, since we use interpolation to create the larger kernels, the number of learned parameters is the same, and initialized from the same kernels for the other datasets. Second, since the number of strides is unchanged, this results is the same number of tokens. Critically, this change has very little effect on the network and its parameters, but enables the model to better capture the information for Charades.

In Table \ref{tab:sota_charades}, we report the results. The core MultiTube approach performs quite well, but with the interpolated kernels, is able to perform on par with TokenLearner \cite{ryoo2021tokenlearner_neurips}, the state-of-the-art, while still sparsely sampling the video. We also perform similarly using significantly less data, e.g., JFT-300M was used to pre-trained TokenLearner, we accomplish the same performance without such large scale data.

\begin{table}
 		\centering
 			\begin{tabular}{lc}
 				\toprule
 				& mAP \\ 
 				\midrule
 				SlowFast \cite{feichtenhofer_iccv_2019} & 	45.2\\
 				AssembleNet-101 \cite{ryoo2019assemblenet} & 	58.6\\
  				AssembleNet++-50 \cite{ryoo2020assemblenet++} & 59.8 \\
  				MoViNet-A6~\cite{kondratyuk2021movinets} & 	63.2 \\
  				TokenLearner \cite{ryoo2021tokenlearner_neurips} & 	66.3 \\
  				\midrule
  				MultiTube Tube-ViT-L & 61.8 \\
  				Interpolated TubeViT-L & 66.2 \\
  				\bottomrule
  			\end{tabular}
  			\caption{Charades classification.}
  			\label{tab:sota_charades}
	\end{table}

\section{Ablations on Tube Shapes.}
In Table \ref{tab:tube_shape}, we provide a detailed study on Kinetics-600 of various tube configurations. We observe that the model isn't overly sensitive to tube shapes, at least on Kinetics-600, but having multiple, different tubes, as well as variation in their shapes is generally beneficial. We use the following tubes in these experiments: 

\begin{enumerate}[label=(\alph*)]
    \item $1\times 16\times 16$
    \item $4\times 32\times 32$
    \item $4\times 4\times 4$
    \item $4\times 12\times 12$
    \item $8\times 8\times 8$
    \item $16\times 4\times 4$
    \item $16\times 16\times 16$
    \item $32\times 8\times 8$
\end{enumerate}

and the following strides:
\begin{enumerate}[label=(\roman*)]
    \item (4, 16, 16)
    \item (8, 8, 8)
    \item (8, 32, 32)
    \item (16, 16, 16)
    \item (32, 32, 32)
\end{enumerate}

\begin{table}[]
    \centering
    \begin{tabular}{c|c}
    \toprule
    Tube Config & K600 \\
    \midrule
    (a+iv) + (b+v) + (f+iv)     &  87.9 \\
    (c+iv) + (e+v) + (g+iv)     & 87.5 \\
    (a+iv) + (e+v) + (g+iv)     & 87.8 \\
    (b+iv) + (e+v) + (g+iv)     & 87.7 \\
    (a+iv) + (d+v) + (e+iv) + (h+v)     &  88.6 \\
    (a+iv) + (b+v) + (c+iv) + (h+v)     &  87.9 \\
    (a+iv) + (d+v) + (e+iv) + (f+v)     &  88.9 \\
    \bottomrule
    \end{tabular}
    \caption{Ablation on different tube shapes, trained for 50k steps.}
    \label{tab:tube_shape}
\end{table}

\clearpage
{\small
\bibliographystyle{ieee_fullname}
\bibliography{egbib}
}

\end{document}